\documentclass{article}
\usepackage{arxiv}
\usepackage[utf8]{inputenc} % allow utf-8 input
\usepackage[T1]{fontenc}    % use 8-bit T1 fonts
\usepackage{hyperref}       % hyperlinks
\usepackage{booktabs}       % professional-quality tables
\usepackage{amsfonts}       % blackboard math symbols
\usepackage{nicefrac}       % compact symbols for 1/2, etc.
\usepackage{microtype}      % microtypography
\usepackage{lipsum}		% Can be removed after putting your text content
\usepackage{graphicx}
\usepackage{doi}
\usepackage{float}
\usepackage{amsmath}
\usepackage{caption}
\usepackage[style=authoryear]{biblatex}  % for author-year citations
\title{Machine Learning for Near Real-Time Identification of Wildfire Burned Areas}

\author{{Khoi Nguyen \hspace{2em} Adam W. Bargteil \hspace{2em} Milton Halem} \\
	Department of Computer Science and Electrical Engineering\\
	University of Maryland, Baltimore County\\
	Baltimore, MD 21250 \\
	\texttt{\{bakhoin1, adamb, halem\}@umbc.edu} \\
}

\date{}

\renewcommand{\headeright}{}
\renewcommand{\undertitle}{}
\renewcommand{\shorttitle}{MACHINE LEARNING FOR IDENTIFYING NEAR REAL-TIME WILDFIRE BURNED AREAS}

\hypersetup{
pdftitle={A template for the arxiv style},
pdfsubject={q-bio.NC, q-bio.QM},
pdfauthor={David S.~Hippocampus, Elias D.~Striatum},
pdfkeywords={First keyword, Second keyword, More},
}

\begin{document}
\maketitle

\begin{abstract}
In this article, we present a near real-time method that estimates burned probability anywhere in the world using satellite images. Our approach takes advantage of the state-of-the-art pix2pix model to translate a preprocessed index calculated from Landsat 8/9 images into an approximate of USGS Burn Probability indicating per pixel burn probability. Since this is a machine learning-based approach, it can be automated with a minimum amount of human intervention with significantly less time delays and geographic restrictions compared to other contemporary methods. We include results from our experiments and compare our approximation with the USGS Burn Probability using RMSE score.

\end{abstract}

% keywords can be removed

\section{Introduction}
The number of extreme wildfires has more than doubled during the past two decades. Wildfires have become larger, more frequent, and more severe. A wildfire can alter an entire ecosystem, cause air pollution and health hazards due to smoke, and destroy millions of lives and properties on its path. It it therefore important to correctly identify burned areas in real-time to provide first responders and fire managers with timely information to contain these mega fires.

A classical approach is to use a Landsat-based index, the delta Normalized Burn Ratio (dNBR) index \cite{key_landscape_2006}. Unfortunately, this classical approach requires manually identifying pre- and post-fire NBR image pairs, which is a difficult task to automate.

More recently, T.J. Hawbaker and colleagues introduced the Burned Probability (BP) products that give per-pixel probability of a location having been burned \parencite{earth_resources_observation_and_science_eros_center_landsat_2022, hawbaker_mapping_2017, hawbaker_landsat_2020}. These products are generated by applying a sophisticated algorithm \parencite{hawbaker_mapping_2017, hawbaker_landsat_2020, todd_j_hawbaker_landsat_2025} to Landsat data, which obviates the need to identify pre-fire images. BP maps are also more granular than dNBR maps as they give probabilistic information per pixel. 

Unfortunately, there are two significant limitations of the USGS Burn Probability maps: time delays and geographic restrictions. In evaluating the rapid spread of a fire, it would take more than 2 weeks for the BP products to be available on the USGS website. Second, BP products are only provided for the Conterminous U.S, thus limiting our scientific capability to study other areas of the globe.

We address these limitations by developing a machine learning algorithm leveraging a model called \textit{pix2pix} \parencite{isola_image--image_2018} to approximate USGS BP products globally in near real-time from Landsat 8 and 9 data. With our trained model, we obtain an approximation output comparable to USGS BP of any location in the world within hours. Our model also serves as a critical component of the NASA Wildfire Digital Twin to provide near real-time validations of wildfire simulations anywhere in the world \parencite{halem_towards_2024}.

%We use Landsat 8 and 9 high-spectral satellite images because they are freely available worldwide. As the current cornerstone satellites of the Landsat Program, Landsat 8 and 9 provide a continuous record of Earth’s land surface. Both Landsat 8 and 9 have 30-meter resolution that render images of size 5000 by 5000 spanning across 11 spectral bands with only 8-day combined revisit period.
%Two sweeps of Landsat data over the August Complex fire in 2020 are used as input. Root Mean Squared Error (RMSE) serves as the metric to compare against the corresponding BP ground truths. Our model approximation demonstrate comparable accuracy to USGS Burn Probability with low RMSE score in the error map.

\section{Background}

\subsection{Delta Normalized Burn Ratio}
Our classical approach estimates real-world burn severity using multiband spectral images captured by Landsat satellites. This method first calculates the Normalized Burn Ratio (NBR) as the difference between the Near Infrared (NIR) and Short Wavelength Infrared (SWIR) channels divided by their sum \parencite{finco_monitoring_2012}. Large NBR values indicate healthy vegetation while small values indicate bare ground or recent fire damage. Next, the delta Normalized Burn Ratio (dNBR) index is computed by subtracting post-fire NBR from pre-fire NBR; higher dNBR values indicate more severe fire damage and lower dNBR values suggest vegetation regrowth.\

\begin{equation}
   \text{NBR} = \frac{\text{Band 5 (NIR)} - \text{Band 7 (SWIR)}}{\text{Band 5 (NIR)} + \text{Band 7 (SWIR)}}
\end{equation}

\begin{equation}
    \text{dNBR} = \text{pre-fire NBR} - \text{post-fire NBR}
\end{equation}

We verified our implementation of dNBR using the Monitoring Trends in Burn Severity (MTBS) dataset \parencite{eidenshink_project_2007} as a reference. The MTBS dataset contains analyst-verified dNBR images of large wildfires over the conterminous U.S, Alaska, and Hawaii from 1984 to 2010. We have successfully reverse engineered the dNBR images as seen in Figure 1. 

Although this approach produces a good estimate of burn severity, it is highly dependent on the selection of pre- and post-fire image pair. To have an accurate evaluation, the images should ideally be cloud-free, smoke-free, and snow-free. The MTBS dataset follows the Initial Assessment approach in \parencite{key_landscape_2006} where the pre-fire image is selected from the same growing season of the previous year, and the post-fire image is collected immediately after the fire is contained before any significant vegetation regrowth. Because the process is labor intensive, this approach is difficult to automate and therefore not ideal for real-time assessment.

\begin{figure}[H]
    \centering
    \includegraphics[width=0.5\textwidth]{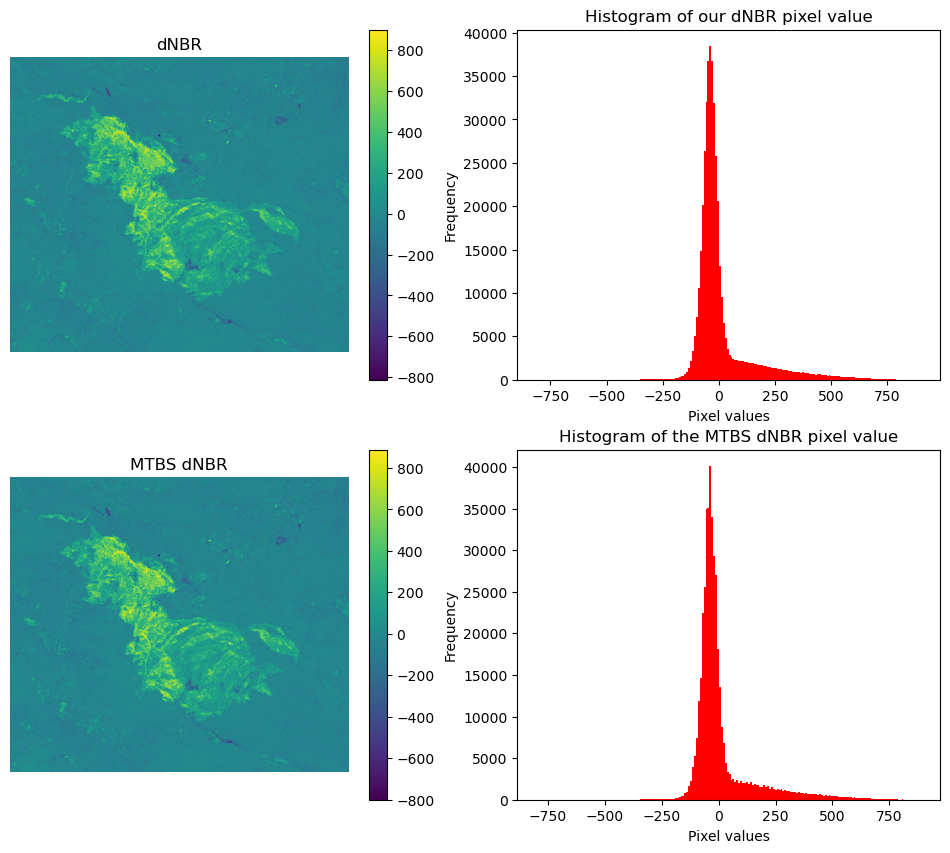}
    \caption{Comparison of MTBS' dNBR and our reverse engineered dNBR. The pixel values distribution of both images are the identical}
    \label{12 landsat sweeps}
\end{figure}

\subsection{Burn Probability}
Burn Probability (BP) is a raster layer of the Landsat Burn Area products produced by USGS that show per-pixel maximum burn probability \parencite{hawbaker_mapping_2017, todd_j_hawbaker_landsat_2025}. As shown in Figure 2 (right), Burn Probability provides a more informative and accurate way to identify and evaluate burned events. In addition, Burn Classification applies a threshold to Burn Probability to render a binary mask indicating burned or unburned areas.

\begin{figure}[H]
    \centering
    \includegraphics[width=0.8\textwidth]{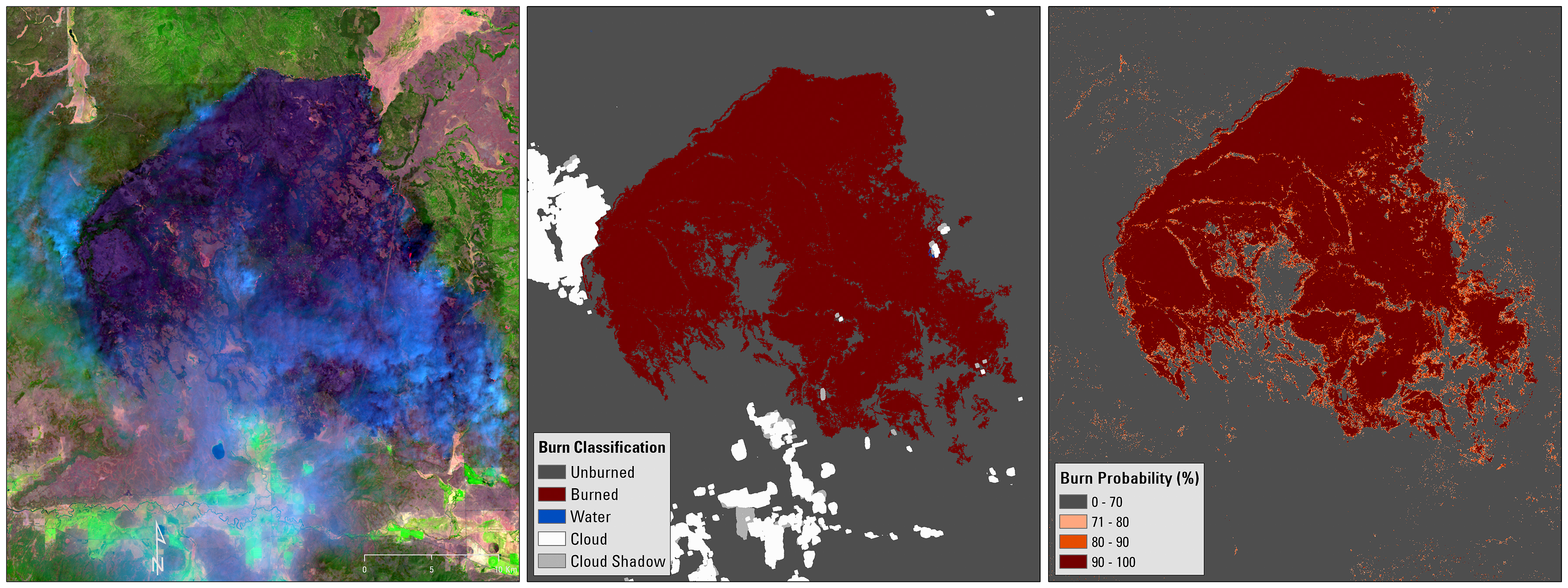}
    \caption{Example of the Landsat Collection 2 Burned Area Science Product showing the Bootleg Fire in Oregon. (Left) Original Landsat Color Map, (Middle) Burn Classification, (Right) Burn Probability
    Source: Landsat Collection 2 Level-3 Burned Area Science Product. }
    \label{12 landsat sweeps}
\end{figure}

BP images are generated by applying the Landsat Burned Area Essential Climate Variable (BAECV) algorithm \parencite{hawbaker_mapping_2017}, which was later replaced by an improved version called the Landsat Burned Area algorithm \parencite{hawbaker_landsat_2020}. The algorithm uses a 3-year window of a scene to estimate the baseline and then calculates the deviation from the baseline. % to check

Unfortunately, there are two significant limitations of the USGS BP maps: time delays and geographic restrictions. Consider the Park Wildfire which raged in California in summer 2024. Starting from 400 acres, it spread to 70,000 acres the next day and to 307,000 the following day consuming an area nearly 500 square miles in 3 days. In evaluating a digital twin simulation of the rapid spread of this fire, it would take 14 to 16 days for the BP products to be available on the USGS website. Moreover, Burn Area products are only provided for the U.S CONUS areas, thus limiting our scientific capability to study other areas of the globe, such as the boreal regions that contribute an alarming share of global wildfires.

\subsection{Image Translation}
With the advancement of AI in recent years, Machine Learning techniques developed in image processing, computer vision, and related fields provide the key to developing the method we seek. Such approaches are well suited to combining the strengths of both the classical method and USGS’s BP products. By training on decades of historical data, we can avoid manually identifying pre- and post-fire images. By building a neural network to estimate Burn Probability, we can compute maps of any location in the world in near real-time - as soon as Landsat data is available .

 We cast our problem as an image translation task where we wish to learn a mapping from an NBR image to its corresponding BP image. \parencite{isola_image--image_2018} presents a general framework for solving image translation problems, called Pix2Pix.

%???? The success of GAN-based architectures \cite{goodfellow_generative_2014} like Pix2Pix lies behind the Game Theory, where the \textbf{Generator} learns to produce realistic output images conditioned on the given input, and the \textbf{Discriminator} simultaneously learns to differentiate between the real output images (the ground truth) and fake images created by the \textbf{Generator}. GANs learn a mapping from an input image to a corresponding output image without the need to hand-engineer loss functions. 

Inspired by Game Theory, Generative Adversarial Networks (GANs) are formulated as minimax games where a Generator is trained adversarially against a Discriminator \parencite{goodfellow_generative_2014}. The \textbf{Generator} learns to produce realistic output images conditioned on the given input while the \textbf{Discriminator} learns to differentiate between the real output images (the ground truth) and fake images created by the \textbf{Generator}. GANs learn a mapping from an input image to a corresponding output image without the need to hand-engineer loss functions.

In pix2pix, the Generator uses a U-net-based network, an encoder-decoder architecture with skip connection used in image segmentation tasks \parencite{ronneberger_u-net_2015}. The skip connections enable both low and high-level information to flow through the network's bottleneck block which helps retain semantic and spatial information as well as the high-resolution details of the image. Unlike other GANs architectures, pix2pix's Discriminator uses a Convolutional PatchGAN that penalizes output at the patch level rather global level, meaning that it classifies each patch as either real or fake and then averages the results to output a global real/fake score. Figure 3 shows the architecture of pix2pix from the original paper \cite{isola_image--image_2018}.

\begin{figure}[H]
    \centering
    \includegraphics[width=0.6\textwidth]{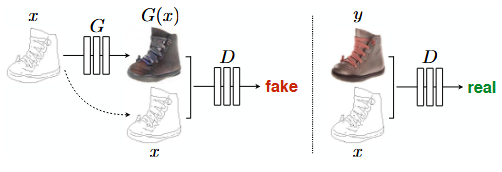}
    \caption{pix2pix architecture from \textit{Image-to-Image Translation with Conditional Adversarial Networks}}
\end{figure}

\subsection{Dataset}
A critical component in any data-driven approach is the choice of dataset. As the current cornerstone satellites of the Landsat Program, Landsat 8 and Landsat 9 \parencite{earth_resources_observation_and_science_eros_center_landsat_2013} provide a continuous record of Earth's land surface since the 1970s. Both Landsat 8 and 9 have 30-meter resolution that render images of size 5000 by 5000 spanning across 11 spectral bands. The Landsat Burn Probability belongs to the Landsat Collection Level-3 Burned Area science product that provides 0 - 100 probability that a location has burned. The data is collected by a script [Hawbaker] that a Machine-to-Machine API download a large volume of data with minimal manual intervention. 

Our dataset consists of matched pairs of Landsat spectral image data and USGS BP products. The input to our network is a Landsat spectral image and the output is an approximation of the BP image. We use the BP images collected from USGS in our loss function during training.

\subsection{Related work}
Burn areas detection have been well studied, from the use of spectral indices derived from satellite images, such as NDVI \parencite{kriegler_preprocessing_1969} and NBR \parencite{key_landscape_2006, epting_evaluation_2005, roy_remote_2006, van_wagtendonk_comparison_2004}, to Machine Learning approaches using classification or segmentation models \parencite{rashkovetsky_wildfire_2021,fan_detecting_2025,padilla_near-real_2025}. Most recently, \parencite{padilla_near-real_2025} developed an algorithm using a neural network to map burn areas globally in near real-time using Sentinel-3 OCLI\&SLSTR reflectance data. Nevertheless, as Sentinel-3 data has coarser spectral resolution (300-meter), the approach is more suitable for regional/global monitoring. 

With 30-meter resolution images from Landsat 8 and 9, our approach offers a richer spectral detail result that can map fire at pixel level, making it effective for area-specific fire detection and assessment.

\section{Wildfire Digital Twin}
There have been ongoing debates around the definition of 'Digital Twin', but in this paper, we require that a Digital Twin must have 4 main entities: Physical Object, Virtual Representation, Real-time Data, and Direct Influence. In the simulations and mathematical models of the past, the virtual world is simply a mirror of its physical counterpart where the data from the physical world is used to update the virtual representation. On the contrary, the Virtual and the Physical World are both first-class citizen in a Digital Twin. The Virtual World assimilates real-time data from sensors to elevate itself to the same status as the real-world Physical Object so the Virtual World can directly influence the Physical Object.  

In that manner, in a Wildfire Digital Twin, the virtual version of a wildfire constantly assimilates the data from LiDAR systems and satellite images to forecast the behavior of the wildfire to provide first-responders with timely information. Actions can then be taken to contain the fire whose status is monitored by the sensors and later used to update the Virtual World.

\section{Methodology}
\label{sec:others}

\subsection{Model}

We cast our problem of estimating BP from Landsat data as an image translation problem and pix2pix \parencite{isola_image--image_2018} performs extremely well on exactly these image translation problems. 

\subsection{Data Preprocessing}
For each Landsat tile in our training dataset, we computed the NBR images over a period of one year. Figure 4 shows 12 NBR images of the same tile in 2020. Then, we calculated the mean (Figure 5) and the standard deviation of those Landsat sweeps (Figure 6). The input NBR images are then normalized by shifting by the mean and scaled by the standard deviation. We refer to these normalized images as nNBR images.
\begin{equation}
    Normalized \, NBR = \frac{Mean - NBR}{Standard \, Deviation}
\end{equation}

Every training pair consists of an nNBR image and its corresponding Burn Probability map. Since these images have considerably high resolution with the size of 5000x5000, we break down each pair of training images into corresponding overlapping tiles of 256x256, so there will be no visible boundaries when we composite those tiles. Then, these pairs will be used to train our model. 

\begin{figure}[h!]
    \centering
    \includegraphics[width=1\textwidth]{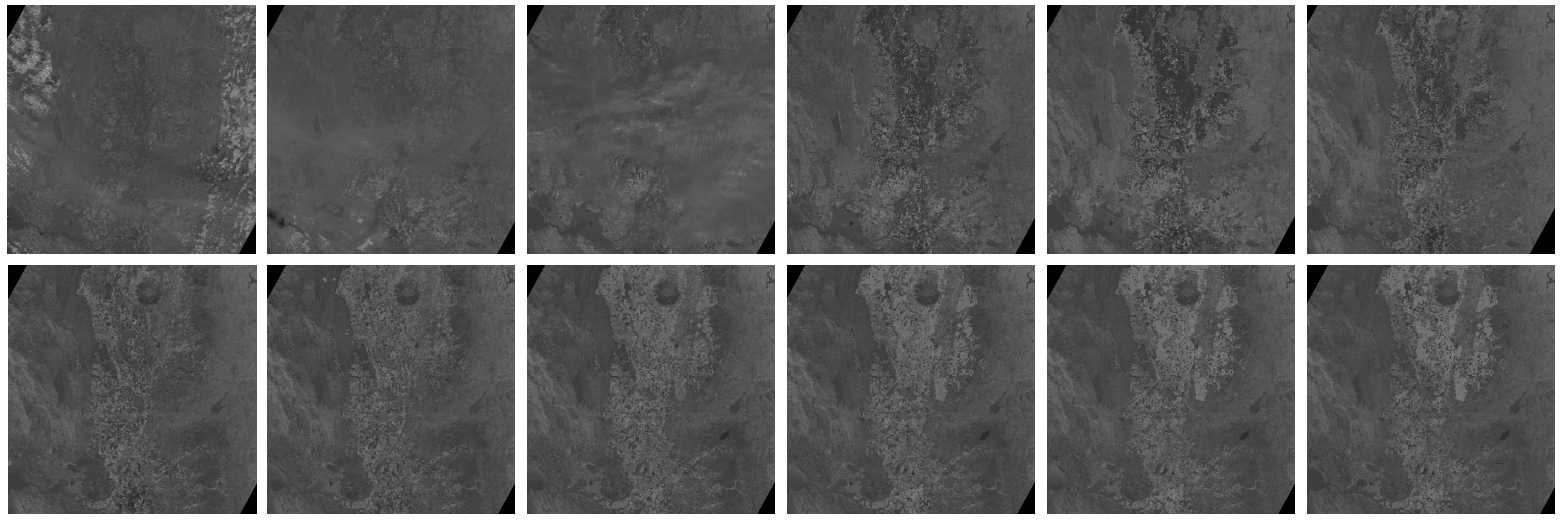}
    \caption{NBR images of the AUGUST COMPLEX wildfire in 2020. These NBR images are processed from the corresponding raw Landsat 8/9 using formula (3)}
    \label{12 landsat sweepss}
\end{figure}

\begin{figure}[h!]
    \centering
    \begin{minipage}{0.48\textwidth}
        \centering
        \includegraphics[width=0.8\textwidth]{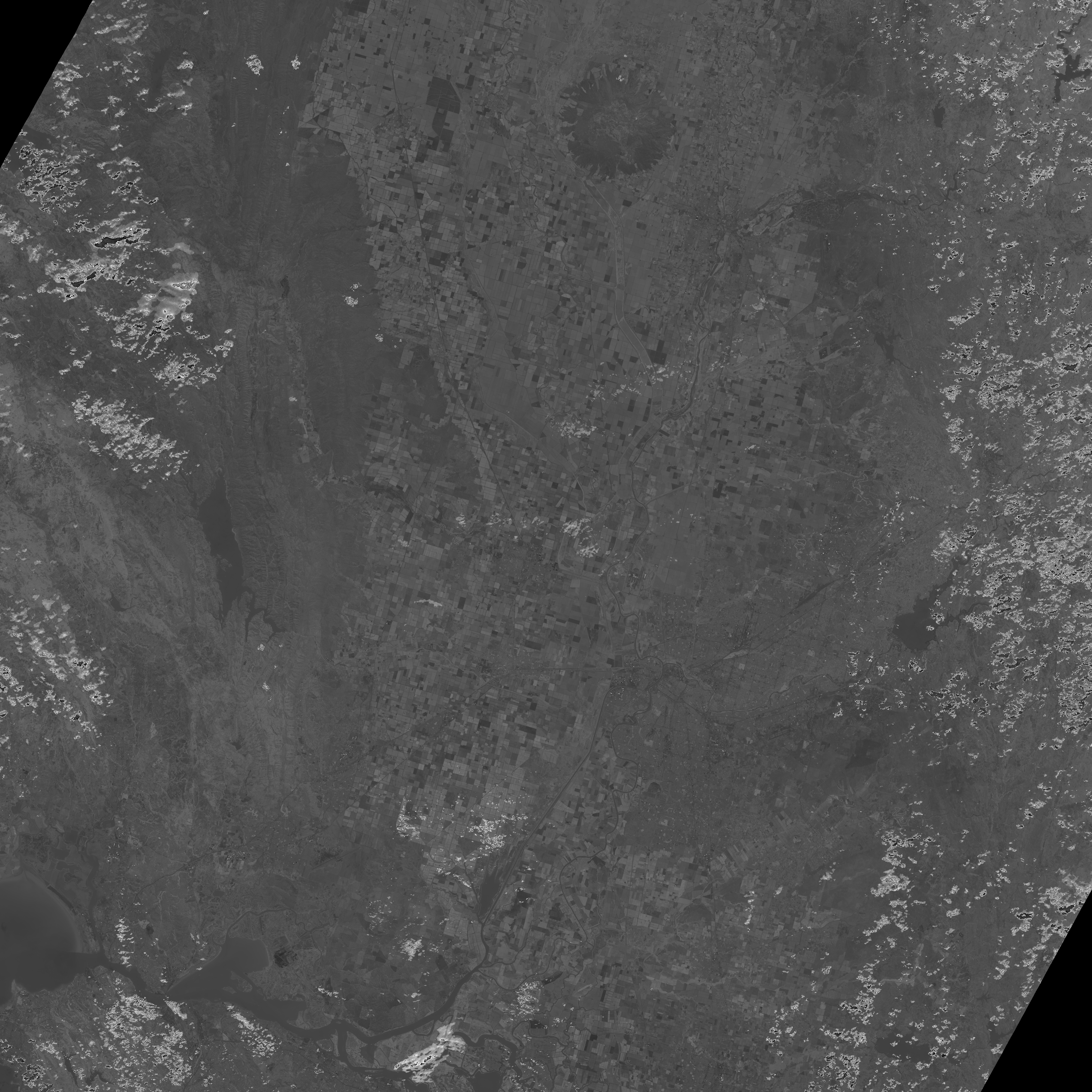}
        \caption{Mean of AUGUST COMPLEX Landsat Sweeps in 2020}
        \label{fig:mean}
    \end{minipage}
    \hfill % adds space between the two images
    \begin{minipage}{0.48\textwidth}
        \centering
        \includegraphics[width=0.8\textwidth]{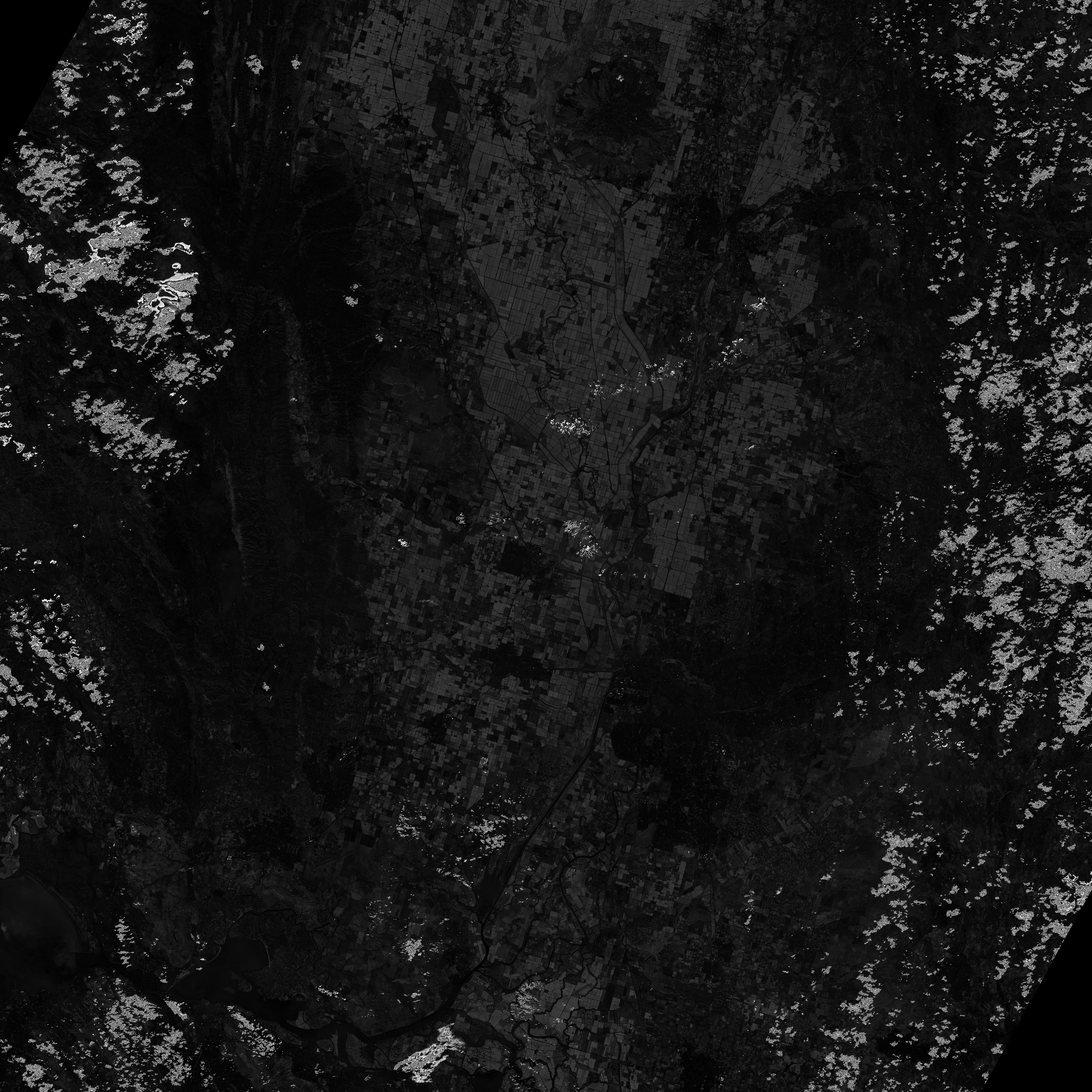}
        \caption{Standard Deviation of Palisades Landsat Sweeps in 2020}
        \label{fig:std}
    \end{minipage}
\end{figure}

\subsection{Compositing images}

% patch edges have less spatial context -> s.a cannot correct systematic errors
%
Our trained pix2pix generator produces overlapping 256x256 tiles that we must composite into the final 5000x5000 output image. Initially, we applied simple averaging of the tile outputs in overlapping regions. Specifically, each pixel $p(i,j)$ in the final output image is calculated as follows:

\begin{equation}
    p(i, j) = \frac{\sum_{k=1}^{N_{ij}} p_k(i,j))}{N_{ij}}
\end{equation}

with $N_{ij}$ as the total number tiles that cover the coordinate $(i,j)$ and $p_k(i,j)$ as the pixel value at (i,j) of the $k^{th}$ patch 

This approach leaves behind artifacts/boundaries in the test set, as illustrated in Figure 8 and 9. For CNN-based models like pix2pix, the predictions near the border are worse than those near the centers where there are more overlapping. Since the errors at the border are biased rather than random, simple averaging the overlapping tiles cannot effectively remove the borders.

\begin{figure}[H]
    \centering
    \includegraphics[width=0.6\textwidth]{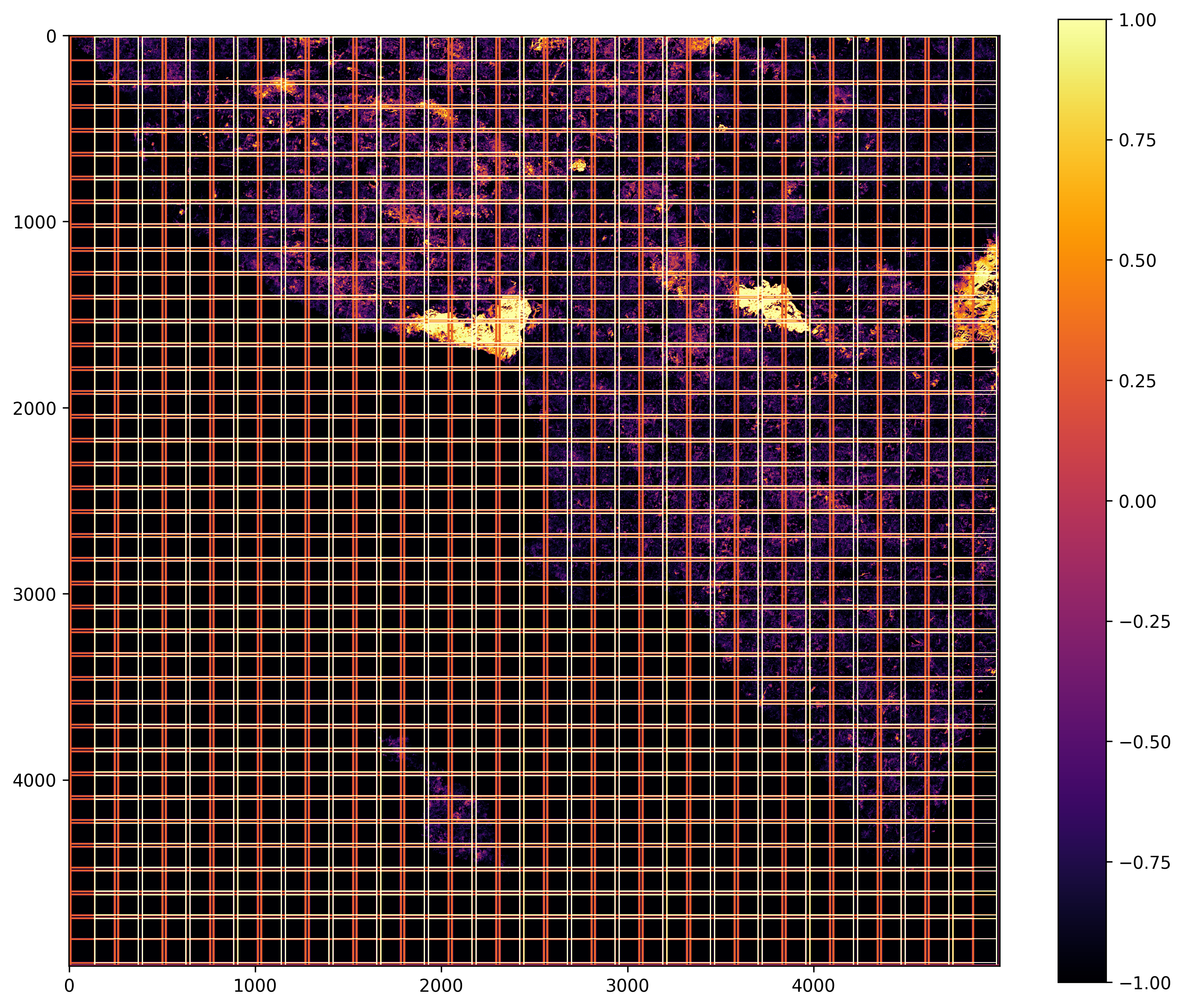}
    \caption{Final Palisades output image composited from overlapping 256x256 tiles with stride=128}
\end{figure}

\begin{figure}[h!]
    \centering
    \begin{minipage}{0.49\textwidth}
        \centering
        \includegraphics[width=0.8\textwidth]{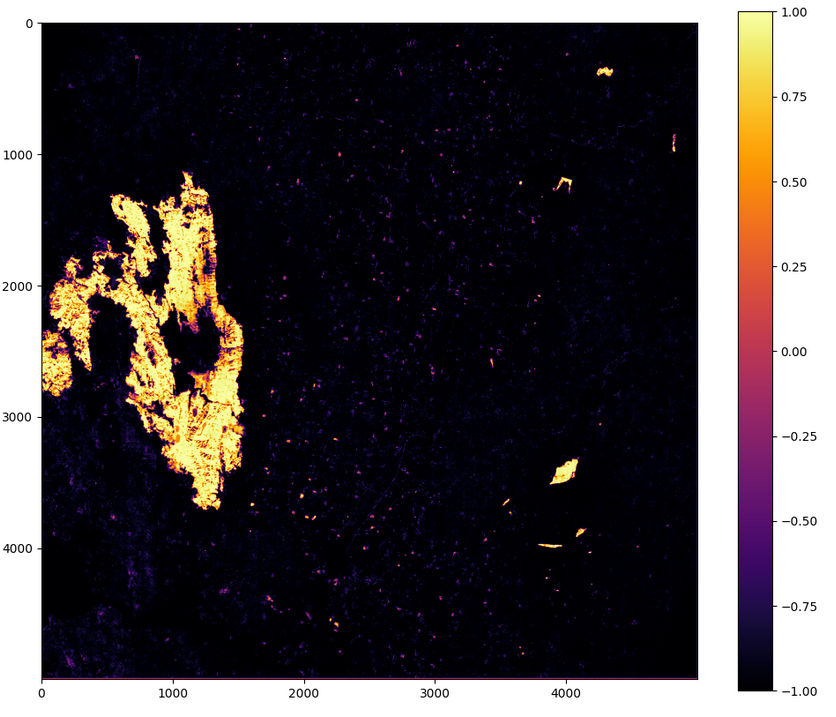}
        \caption{AUGUST Complex - Simple Averaging}
        \label{fig:mean}
    \end{minipage}
    \hfill % adds space between the two images
    \begin{minipage}{0.49\textwidth}
        \centering
        \includegraphics[width=0.8\textwidth]{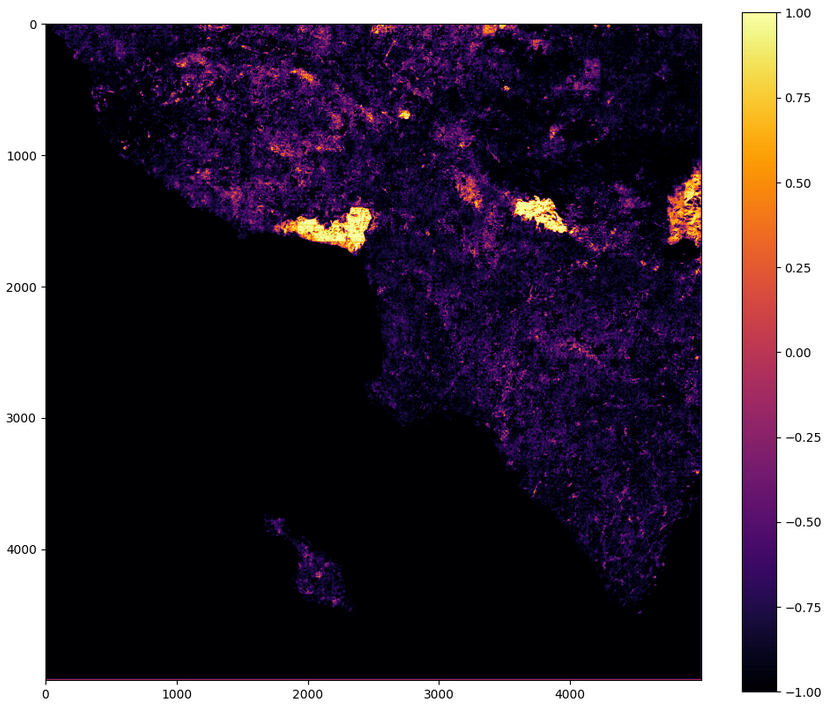}
        \caption{Palisades - Simple Averaging}
        \label{fig:std}
    \end{minipage}
\end{figure}

With, this simple approach artifacts remain. We achieved better results using weighted blending, 
\begin{equation} 
    p(i, j) = \frac{\sum_{k=1}^{N_{ij}}(w_k(i,j)\cdot p_k(i,j))}{\sum_{k=1}^{N_{ij}}w_k(i,j)}.
\end{equation}

We experimented with bilinear,
\begin{equation} 
    w(i, j) = \left(1 - \left|\frac{i}{W/2}\right|\right) \times \left(1 - \left|\frac{j}{H/2}\right|\right),
\end{equation}
and Gaussian weights,
\begin{equation} 
    w(i, j) = A \exp \left( -\left( \frac{(i - i_0)^2}{2\sigma_i^2} + \frac{(j - j_0)^2}{2\sigma_j^2} \right) \right).
\end{equation}

Both bilinear (Figure 10, 11) and Gaussian blending (Figure 12, 13) give similar results and perform better than simple averaging. Because we truncated the Gaussian, the running times were similar as well.

%Instead of averaging all the pixels of their contribution, we used bilinear blending where each pixel is assigned a weight with regards to their position to the center; the closer the pixel is to the center, the higher weight it's assigned and vice versa: This means that pixels near the edges have the lowest weights and contribute the least to the final output. 

%In addition to bilinear blending, we tried using 2D Gaussian function to assign weights and smooth out the seams. Unlike bilinear weights that decrease linearly as the distance from the center increases, 2D Gaussian weights decreases more smoothly from the center due to assumption of the Gaussian (or Normal) distribution of pixel weights: The weighs are accumulated and then used to normalize the overlapping regions

%We calculate every pixel in the final output for both Bilinear Blending and Gaussian Smoothing as follows:

\begin{figure}[H]
    \centering
    \begin{minipage}{0.49\textwidth}
        \centering
        \includegraphics[width=0.8\textwidth]{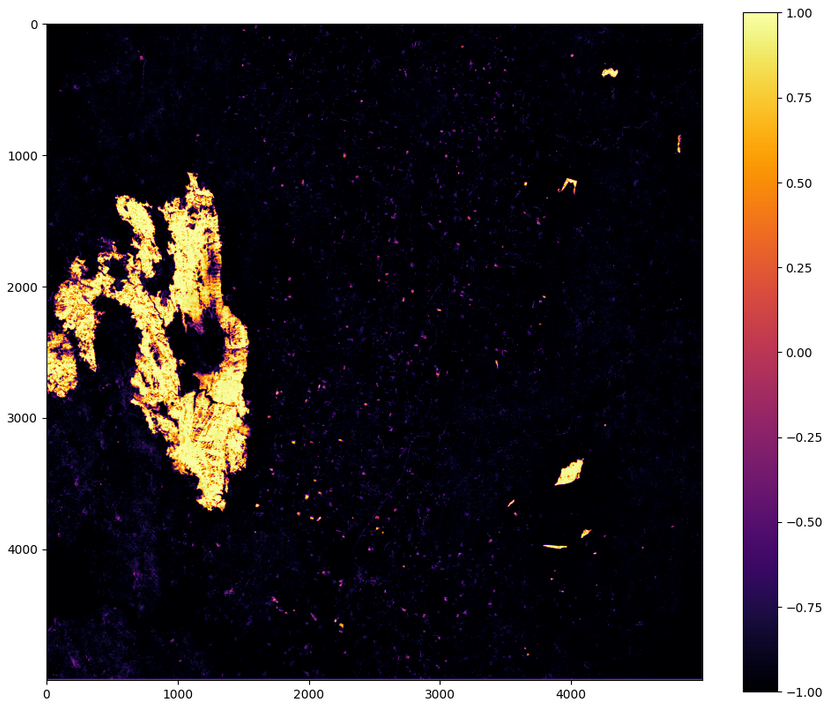}
        \caption{AUGUST Complex - Bilinear Blending}
        \label{fig:mean}
    \end{minipage}
    \hfill % adds space between the two images
    \begin{minipage}{0.49\textwidth}
        \centering
        \includegraphics[width=0.8\textwidth]{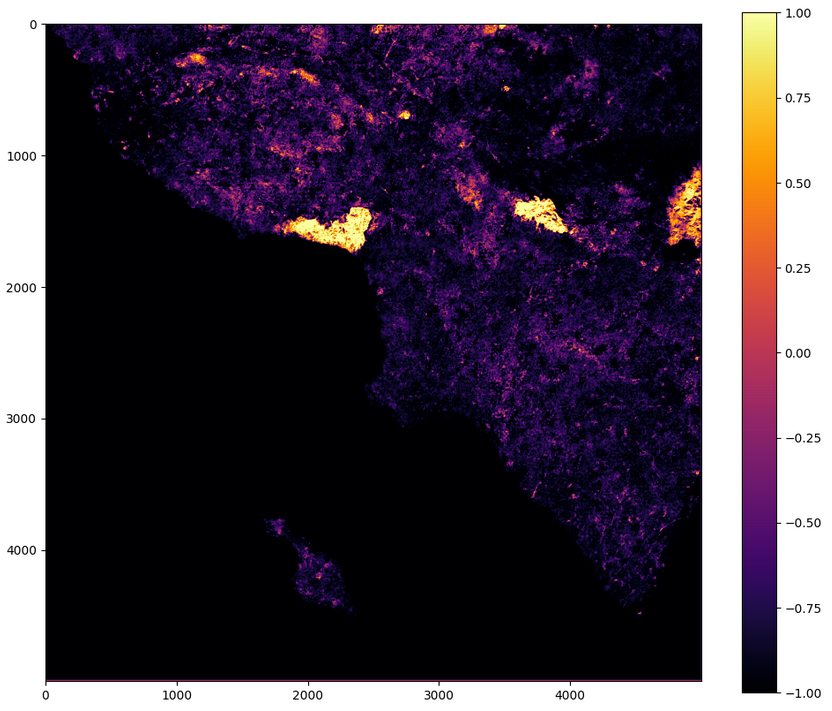}
        \caption{Palisades - Bilinear Blending}
        \label{fig:std}
    \end{minipage}
\end{figure}

\begin{figure}[H]
    \centering
    \begin{minipage}{0.49\textwidth}
        \centering
        \includegraphics[width=0.8\textwidth]{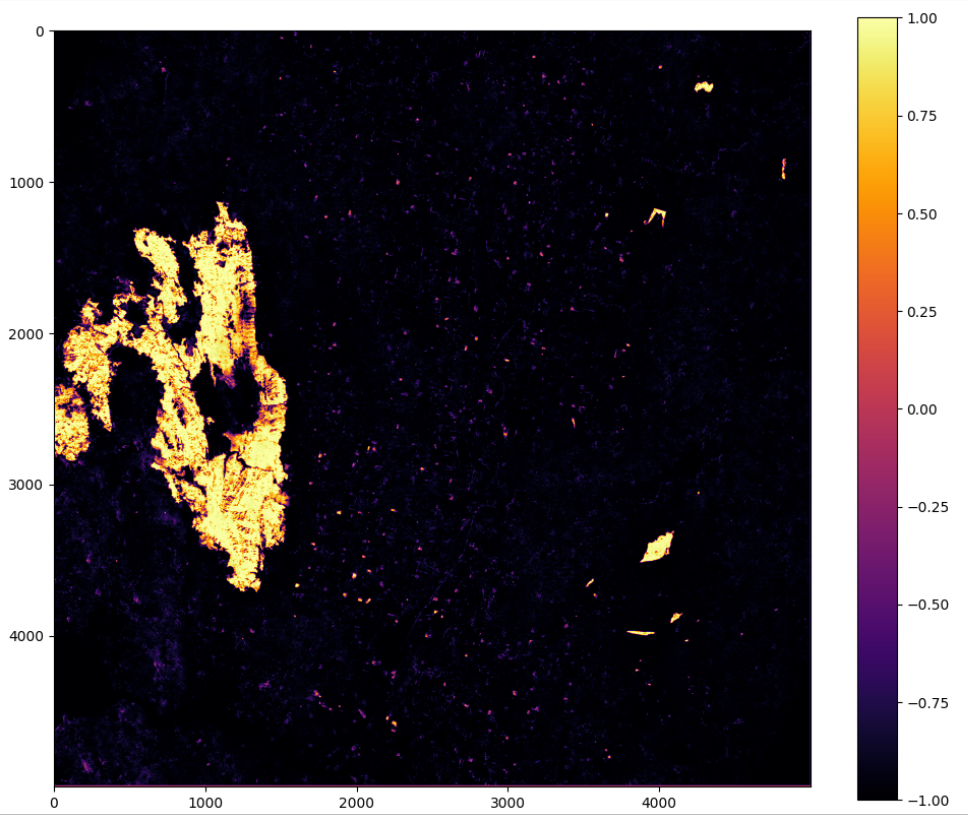}
        \caption{AUGUST Complex - 2D Gaussian Smoothing}
        \label{fig:mean}
    \end{minipage}
    \hfill
    \begin{minipage}{0.49\textwidth}
        \centering
        \includegraphics[width=0.8\textwidth]{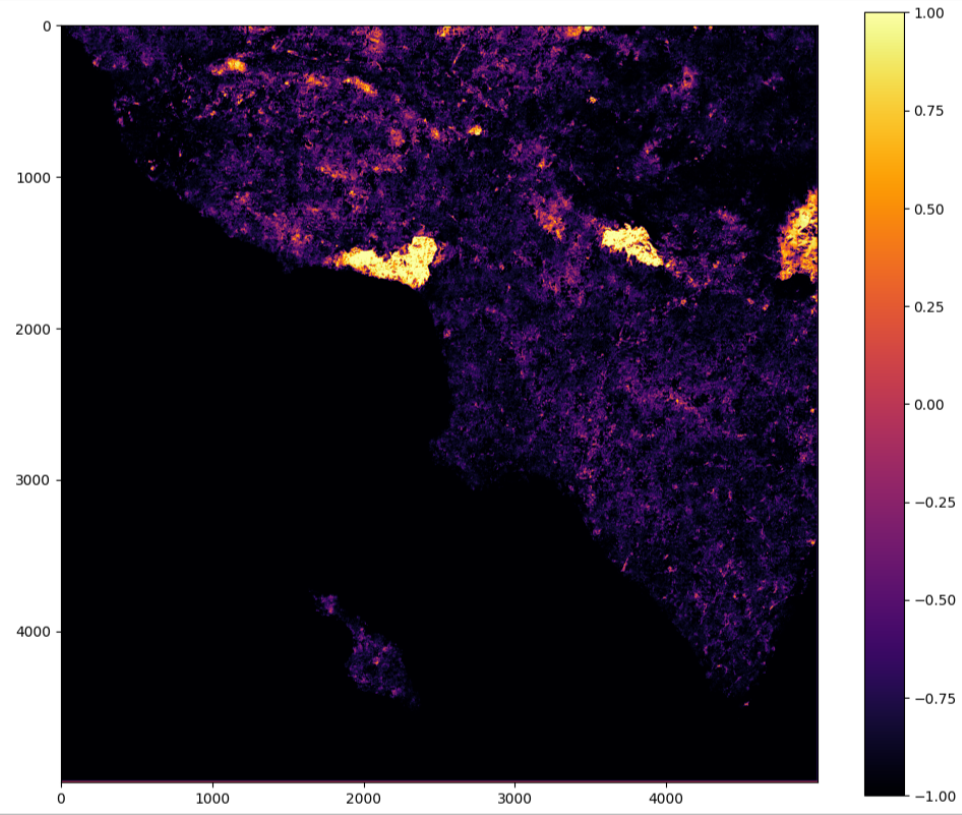}
        \caption{Palisades - Gaussian Blending}
        \label{fig:std}
    \end{minipage}
\end{figure}

Though both Gaussian Smoothing and Bilinear Blending gives good and similar results, one might choose Bilinear Blending over Gaussian Smoothing. Since 2D Gaussian extends infinitely to all directions and is non-zero everywhere, we only calculated Gaussian weights for the pixels inside the kernel and truncated what is outside. Consequently, Gaussian Smoothing might take longer to render a composite image than does Bilinear Blending, whose weights decreases linearly to 0, and might leave behind visible artifacts.

\section{Results}
We first tested our model on a different August Complex sweep (Figure 14). We mapped out the water bodies in the BP map to focus on the burned areas.

\begin{figure}[H]
    \centering
    \includegraphics[width=1\textwidth]{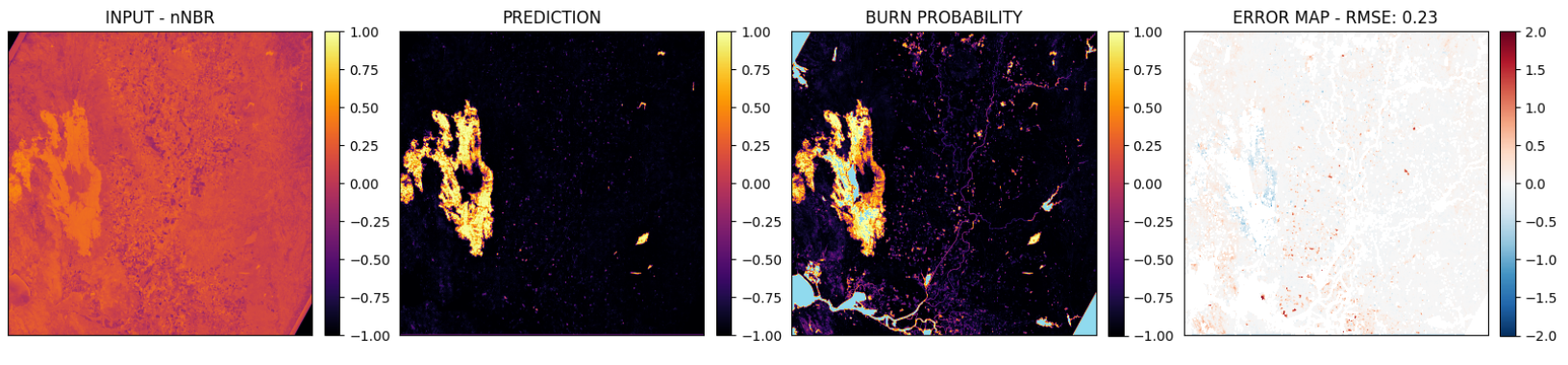}
    \caption{Model's output for August Complex wildfire. Burn Probability (ground truth) uses a blue mask to mask out water bodies. The Prediction values ranging from -1.0 to 1.0, the higher the values, the more likely that the pixel indicates a burned area}
    \label{Output 1}
\end{figure}

We then tested the model on the Palisades wildfire scene in our test set (Figure 15). The model does not perform as well on this test image since cloud and smoke obscure the ground. % to ask: what else should I write here?

\begin{figure}[H]
    \centering
    \includegraphics[width=1\textwidth]{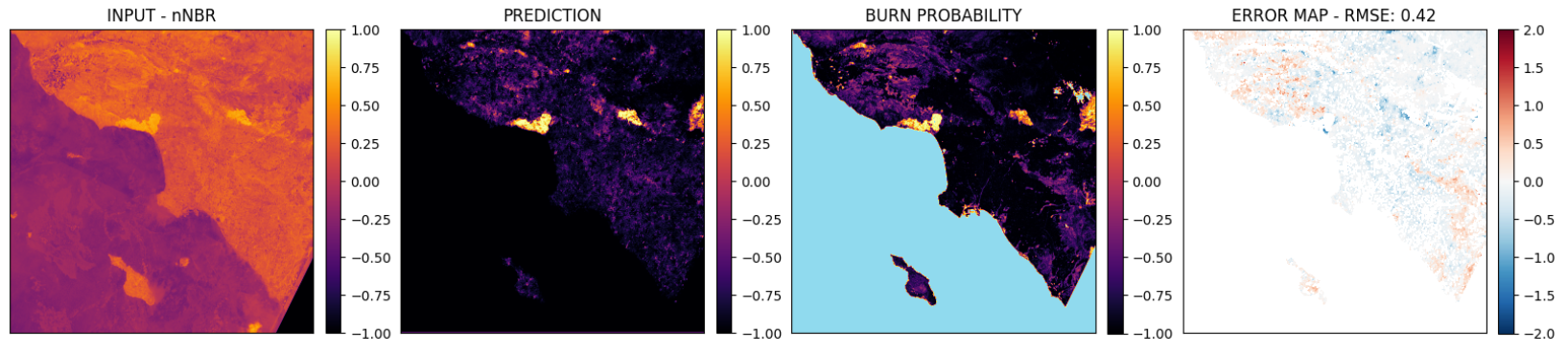}
    \caption{Model's output for The Palisades wildfire. Burn Probability (ground truth) uses a blue mask to mask out water bodies. The Prediction values ranging from -1.0 to 1.0, the higher the values, the more likely that the pixel indicates a burned area}
    \label{Output 2}
\end{figure}

\section{Future Work}
% color channels to map out water
While our pipeline is able to take an individual Landsat image, compute nNBR and then use pix2pix to produce burn probability, we currently must mask out clouds, water, and other problematic pixels before processing. A fully automated operational system would need to automatically mask out these pixels. This task could be accomplished as a pre-process using the existing techniques employed by USGS, but another possible avenue for future work would be to use the additional color channels supported by pix2pix, which was developed to handle 3-channel RGB images, to automatically detect water and clouds concurrently with estimating burn probability.

% robust statistics
To better approximate the Burn Probability maps, one could use robust statistics to improve the nNBR images. Since the nNBR is scaled by shifting the mean and scaled by the standard deviation, it is highly sensitive to extreme values, which in our case includes water bodies or clouds. Robust metrics like median and median absolute deviation can help mitigate the effect of these extreme values as they are more resilient to outliers. One caveat to this method is computational efficiency because of the need to sort all the corresponding pixels of different sweeps of a particular scene. 

Though bilinear blending and gaussian blending give good results, future work could explore using Graph Cut \parencite{boykov_fast_2001} to composite the tiles. Instead of averaging the pixels, Graph Cut techniques choose pixel values from among the overlapping tiles to minimize visual discontinuities. In this way, each pixel value in the final image appears in one of the tiles. \parencite{agarwala_interactive_2004} combines Graph Cut with Gradient-Domain fusion to further minimize visual discontinuities while preserving the gradients of the tiles rather than the absolute values. 

%Instead of averaging the pixels, Graph Cut techniques choose the best seam so that the difference between the patches is minimal. By doing this, we avoid interfering the model's original output.  combines Graph Cut with Gradient-Domain fusion to choose the best seam and remove artifacts after joining the images. 

\section{Acknowledgement}
% Refer to Dr. Hawbaker's assistance
Special thanks to Dr. T.J. Hawbaker for sharing his expertise in Burn Probability mapping and for his assistance in acquiring Landsat 8/9 dataset. 

This research was funded by the National Aeronautics and Space Administration - NASA (Grant No.12345). The computing resources provided by the UMBC High Performance Computing Facility were critical to the experiments in this study. 

%\bibliography{references} %%% Uncomment this line and comment out the ``thebibliography'' section below to use the external .bib file (using bibtex) .
\printbibliography

@article{hawbaker_mapping_2017,
	title = {Mapping burned areas using dense time-series of Landsat data},
	volume = {198},
	issn = {0034-4257},
	url = {https://www.sciencedirect.com/science/article/pii/S0034425717302857},
	doi = {10.1016/j.rse.2017.06.027},
	pages = {504--522},
	journaltitle = {Remote Sensing of Environment},
	shortjournal = {Remote Sensing of Environment},
	author = {Hawbaker, Todd J. and Vanderhoof, Melanie K. and Beal, Yen-Ju and Takacs, Joshua D. and Schmidt, Gail L. and Falgout, Jeff T. and Williams, Brad and Fairaux, Nicole M. and Caldwell, Megan K. and Picotte, Joshua J. and Howard, Stephen M. and Stitt, Susan and Dwyer, John L.},
	urldate = {2025-11-09},
	date = {2017-09-01},
}

@inproceedings{halem_towards_2024,
	location = {Cham},
	title = {Towards a Dynamic Data Driven Wildfire Digital Twin ({WDT}): Impacts on Deforestation, Air Quality and Cardiopulmonary Disease},
	isbn = {978-3-031-52670-1},
	pages = {403--410},
	booktitle = {Dynamic Data Driven Applications Systems},
	publisher = {Springer Nature Switzerland},
	author = {Halem, M. and Kochanski, A. K. and Mandel, J. and Sleeman, J. and Demoz, B. and Bargteil, A. and Patil, S. and {S.Shivadekar} and Iorga, A. and Dorband, J. and Mackinnon, J. and Chiao, S. and Yang, Z. and Yesha, Ya. and Sorkin, J. and Kalnay, E. and Safa, S. and Da, C.},
	editor = {Blasch, Erik and Darema, Frederica and Aved, Alex},
	date = {2024},
}

@misc{earth_resources_observation_and_science_eros_center_landsat_2022,
	title = {Landsat Level-3 Burned Area, Collection 2},
	rights = {Creative Commons Zero v1.0 Universal},
	url = {https://www.usgs.gov/centers/eros/science/usgs-eros-archive-landsat-collection-2-level-3-burned-area-ba-science-product},
	doi = {10.5066/P999KPGS},
	publisher = {U.S. Geological Survey},
	author = {{Earth Resources Observation and Science (EROS) Center}},
	urldate = {2025-11-10},
	date = {2022},
}

@misc{todd_j_hawbaker_landsat_2025,
	title = {The Landsat Collection 2 Burned Area Products for the conterminous United States (ver. 2.0, April 2024)},
	rights = {Creative Commons Zero v1.0 Universal},
	url = {https://www.sciencebase.gov/catalog/item/63f101f3d34efa0476b03d64},
	doi = {10.5066/P9F26LY6},
	publisher = {U.S. Geological Survey},
	author = {Todd J Hawbaker and Melanie K Vanderhoof and Casey E. Menick and Gail L Schmidt and Joshua J Picotte},
	urldate = {2025-11-10},
	date = {2025-03-14},
}

@article{hawbaker_landsat_2020,
	title = {The Landsat Burned Area algorithm and products for the conterminous United States},
	volume = {244},
	issn = {00344257},
	url = {https://linkinghub.elsevier.com/retrieve/pii/S0034425720301711},
	doi = {10.1016/j.rse.2020.111801},
	pages = {111801},
	journaltitle = {Remote Sensing of Environment},
	shortjournal = {Remote Sensing of Environment},
	author = {Hawbaker, Todd J. and Vanderhoof, Melanie K. and Schmidt, Gail L. and Beal, Yen-Ju and Picotte, Joshua J. and Takacs, Joshua D. and Falgout, Jeff T. and Dwyer, John L.},
	urldate = {2025-11-10},
	date = {2020-07},
	langid = {english},
}

@misc{isola_image--image_2018,
	title = {Image-to-Image Translation with Conditional Adversarial Networks},
	url = {http://arxiv.org/abs/1611.07004},
	doi = {10.48550/arXiv.1611.07004},
	number = {{arXiv}:1611.07004},
	publisher = {{arXiv}},
	author = {Isola, Phillip and Zhu, Jun-Yan and Zhou, Tinghui and Efros, Alexei A.},
	urldate = {2025-11-10},
	date = {2018-11-26},
	eprinttype = {arxiv},
	eprint = {1611.07004 [cs]},
}

@article{finco_monitoring_2012,
	title = {Monitoring Trends and Burn Severity ({MTBS}): Monitoring wildfire activity for the past quarter century using landsat data},
	author = {Finco, Mark},
	date = {2012},
	langid = {english},
}

@article{key_landscape_2006,
	title = {Landscape Assessment ({LA})},
	author = {Key, Carl H and Benson, Nathan C},
	date = {2006},
	langid = {english},
}

@misc{ronneberger_u-net_2015,
	title = {U-Net: Convolutional Networks for Biomedical Image Segmentation},
	url = {http://arxiv.org/abs/1505.04597},
	doi = {10.48550/arXiv.1505.04597},
	shorttitle = {U-Net},
	number = {{arXiv}:1505.04597},
	publisher = {{arXiv}},
	author = {Ronneberger, Olaf and Fischer, Philipp and Brox, Thomas},
	urldate = {2025-11-14},
	date = {2015-05-18},
	eprinttype = {arxiv},
	eprint = {1505.04597 [cs]},
}

@article{eidenshink_project_2007,
	title = {A Project for Monitoring Trends in Burn Severity},
	volume = {3},
	issn = {1933-9747},
	url = {https://fireecology.springeropen.com/articles/10.4996/fireecology.0301003},
	doi = {10.4996/fireecology.0301003},
	pages = {3--21},
	number = {1},
	journaltitle = {Fire Ecology},
	shortjournal = {fire ecol},
	author = {Eidenshink, Jeff and Schwind, Brian and Brewer, Ken and Zhu, Zhi-Liang and Quayle, Brad and Howard, Stephen},
	urldate = {2025-11-18},
	date = {2007-06},
	langid = {english},
}

@misc{goodfellow_generative_2014,
	title = {Generative Adversarial Networks},
	url = {http://arxiv.org/abs/1406.2661},
	doi = {10.48550/arXiv.1406.2661},
	number = {{arXiv}:1406.2661},
	publisher = {{arXiv}},
	author = {Goodfellow, Ian J. and Pouget-Abadie, Jean and Mirza, Mehdi and Xu, Bing and Warde-Farley, David and Ozair, Sherjil and Courville, Aaron and Bengio, Yoshua},
	urldate = {2025-11-18},
	date = {2014-06-10},
	eprinttype = {arxiv},
	eprint = {1406.2661 [stat]},
}

@misc{earth_resources_observation_and_science_eros_center_landsat_2013,
	title = {Landsat 8-9 Operational Land Imager / Thermal Infrared Sensor Level-1, Collection 2},
	rights = {Creative Commons Zero v1.0 Universal},
	url = {https://www.usgs.gov/centers/eros/science/usgs-eros-archive-landsat-archives-landsat-8-9-operational-land-imager-and},
	doi = {10.5066/P975CC9B},
	publisher = {U.S. Geological Survey},
	author = {{Earth Resources Observation and Science (EROS) Center}},
	urldate = {2025-11-18},
	date = {2013},
}

@article{padilla_near-real_2025,
	title = {Near-real time monitoring of burned area at global scale based on deep learning},
	volume = {46},
	issn = {0143-1161, 1366-5901},
	url = {https://www.tandfonline.com/doi/full/10.1080/01431161.2025.2528256},
	doi = {10.1080/01431161.2025.2528256},
	pages = {5996--6038},
	number = {16},
	journaltitle = {International Journal of Remote Sensing},
	shortjournal = {International Journal of Remote Sensing},
	author = {Padilla, Marc and Ramo, Ruben and Gomez-Dans, Jose Luis and Sierra, Sergio and Mota, Bernardo and Lacaze, Roselyne and Tansey, Kevin},
	urldate = {2025-11-27},
	date = {2025-08-18},
	langid = {english},
}

@article{rashkovetsky_wildfire_2021,
	title = {Wildfire Detection From Multisensor Satellite Imagery Using Deep Semantic Segmentation},
	volume = {14},
	rights = {https://creativecommons.org/licenses/by/4.0/legalcode},
	issn = {1939-1404, 2151-1535},
	url = {https://ieeexplore.ieee.org/document/9468971/},
	doi = {10.1109/JSTARS.2021.3093625},
	pages = {7001--7016},
	journaltitle = {{IEEE} Journal of Selected Topics in Applied Earth Observations and Remote Sensing},
	shortjournal = {{IEEE} J. Sel. Top. Appl. Earth Observations Remote Sensing},
	author = {Rashkovetsky, Dmitry and Mauracher, Florian and Langer, Martin and Schmitt, Michael},
	urldate = {2025-11-27},
	date = {2021},
}

@article{roy_remote_2006,
	title = {Remote Sensing of Fire Severity: Assessing the Performance of the Normalized Burn Ratio},
	volume = {3},
	rights = {https://ieeexplore.ieee.org/Xplorehelp/downloads/license-information/{IEEE}.html},
	issn = {1545-598X},
	url = {http://ieeexplore.ieee.org/document/1576701/},
	doi = {10.1109/LGRS.2005.858485},
	shorttitle = {Remote Sensing of Fire Severity},
	pages = {112--116},
	number = {1},
	journaltitle = {{IEEE} Geoscience and Remote Sensing Letters},
	shortjournal = {{IEEE} Geosci. Remote Sensing Lett.},
	author = {Roy, D.P. and Boschetti, L. and Trigg, S.N.},
	urldate = {2025-11-27},
	date = {2006-01},
	langid = {english},
}

@article{epting_evaluation_2005,
	title = {Evaluation of remotely sensed indices for assessing burn severity in interior Alaska using Landsat {TM} and {ETM}+},
	volume = {96},
	issn = {0034-4257},
	url = {https://www.sciencedirect.com/science/article/pii/S0034425705001021},
	doi = {10.1016/j.rse.2005.03.002},
	pages = {328--339},
	number = {3},
	journaltitle = {Remote Sensing of Environment},
	shortjournal = {Remote Sensing of Environment},
	author = {Epting, Justin and Verbyla, David and Sorbel, Brian},
	urldate = {2025-12-06},
	date = {2005-06-30},
}

@article{van_wagtendonk_comparison_2004,
	title = {Comparison of {AVIRIS} and Landsat {ETM}+ detection capabilities for burn severity},
	volume = {92},
	rights = {https://www.elsevier.com/tdm/userlicense/1.0/},
	issn = {00344257},
	url = {https://linkinghub.elsevier.com/retrieve/pii/S003442570400152X},
	doi = {10.1016/j.rse.2003.12.015},
	pages = {397--408},
	number = {3},
	journaltitle = {Remote Sensing of Environment},
	shortjournal = {Remote Sensing of Environment},
	author = {Van Wagtendonk, Jan W. and Root, Ralph R. and Key, Carl H.},
	urldate = {2025-12-06},
	date = {2004-08},
	langid = {english},
}

@article{fan_detecting_2025,
	title = {Detecting Burned Vegetation Areas by Merging Spectral and Texture Features in a {ResNet} Deep Learning Architecture},
	volume = {17},
	rights = {http://creativecommons.org/licenses/by/3.0/},
	issn = {2072-4292},
	url = {https://www.mdpi.com/2072-4292/17/22/3665},
	doi = {10.3390/rs17223665},
	number = {22},
	journaltitle = {Remote Sensing},
	author = {Fan, Jiahui and Yao, Yunjun and Li, Yajie and Zhang, Xueyi and Chen, Jiquan and Fisher, Joshua B. and Zhang, Xiaotong and Jiang, Bo and Liu, Lu and Xie, Zijing and Zhang, Luna and Qiu, Fei and Fan, Jiahui and Yao, Yunjun and Li, Yajie and Zhang, Xueyi and Chen, Jiquan and Fisher, Joshua B. and Zhang, Xiaotong and Jiang, Bo and Liu, Lu and Xie, Zijing and Zhang, Luna and Qiu, Fei},
	urldate = {2025-12-07},
	date = {2025-11-06},
	langid = {english},
	note = {Company: Multidisciplinary Digital Publishing Institute
Distributor: Multidisciplinary Digital Publishing Institute
Institution: Multidisciplinary Digital Publishing Institute
Label: Multidisciplinary Digital Publishing Institute
Publisher: publisher},
}

@article{kriegler_preprocessing_1969,
	title = {Preprocessing transformations and their effect on multispectral recognition},
	volume = {6},
	pages = {97--132},
	author = {Kriegler, F.J. and Malila, W.A. and Nalepka, R.F. and Richardson, .},
	date = {1969},
}

@article{boykov_fast_2001,
	title = {Fast approximate energy minimization via graph cuts},
	volume = {23},
	rights = {https://ieeexplore.ieee.org/Xplorehelp/downloads/license-information/{IEEE}.html},
	issn = {01628828},
	url = {http://ieeexplore.ieee.org/document/969114/},
	doi = {10.1109/34.969114},
	pages = {1222--1239},
	number = {11},
	journaltitle = {{IEEE} Transactions on Pattern Analysis and Machine Intelligence},
	shortjournal = {{IEEE} Trans. Pattern Anal. Machine Intell.},
	author = {Boykov, Y. and Veksler, O. and Zabih, R.},
	urldate = {2026-02-06},
	date = {2001-11},
}

@article{agarwala_interactive_2004,
	title = {Interactive digital photomontage},
	volume = {23},
	issn = {0730-0301, 1557-7368},
	url = {https://dl.acm.org/doi/10.1145/1015706.1015718},
	doi = {10.1145/1015706.1015718},
	pages = {294--302},
	number = {3},
	journaltitle = {{ACM} Transactions on Graphics},
	shortjournal = {{ACM} Trans. Graph.},
	author = {Agarwala, Aseem and Dontcheva, Mira and Agrawala, Maneesh and Drucker, Steven and Colburn, Alex and Curless, Brian and Salesin, David and Cohen, Michael},
	urldate = {2026-02-06},
	date = {2004-08},
	langid = {english},
}





% \end{thebibliography}

\end{document}